%% file: main.tex
\documentclass[runningheads]{llncs}
\usepackage[T1]{fontenc}
\usepackage{graphicx}
\usepackage{booktabs} 
\usepackage{amsmath}
\usepackage{amssymb}
\usepackage{hyperref}
\usepackage{color}
\usepackage{algorithm}
\usepackage{algpseudocode}

\begin{document}
\title{Towards Inclusive Mobility Modeling: Characterizing and Evaluating Elderly Trajectory Patterns in Urban Systems\thanks{To appear in 7th International Conference on Social Computing (ICSC)}}

%
%
\author{Zhengxuan Wang\inst{1,2}$^{\star}$, 
Haohan He\inst{3}$^{\star}$,
Mengying Zhou\inst{1,2}$^{\dagger}$}
\authorrunning{Wang et al.}
%
\institute{
School of Computing and Artificial Intelligence, Shanghai University of Finance and Economics, Shanghai, China
\and
MOE Key Laboratory of Interdisciplinary Research of Computation and Economics, Shanghai University of Finance and Economics, Shanghai, China
\\
\and
Center for Data Science, New York University, New York, USA
\email{zhoumengying@sufe.edu.cn}
}
\maketitle              

\begingroup
\renewcommand{\thefootnote}{\relax} 
\footnotetext{$^{\star}$Equal contribution.}
\footnotetext{$^{\dagger}$Corresponding author.}
\endgroup

\input{0_abstract}
\input{1_introduction}
\input{2_related_work}
\input{3_analysis}
\input{4_generative_experiments}
\input{5_conclusions_and_policy_implications}

%
%

\bibliographystyle{splncs04}
\bibliography{mybibliography}

\end{document}

%% file: 0_abstract.tex
\begin{abstract}
The rapid advance of smart cities increasingly depends on trajectory data mining, yet underrepresented demographic groups, particularly the elderly, are often sparsely represented in public mobility datasets. This underrepresentation can introduce systematic bias into mobility modeling and downstream urban planning. Using the 2016--2020 Jersey City subset of the Citi Bike System Data, this study quantitatively examines how the absence of underrepresented subgroups’ mobility signatures affects mobility modeling, using synthetic trajectory generation as a case study. The analysis reveals that elderly riders exhibit a structurally distinct mobility signature, including localized activity spaces (958 m vs.\ 1,189 m for young riders), lower mobility entropy (1.82 vs.\ 4.15), and asymmetric off-peak temporal patterns. To demonstrate that relying on majority-dominated training data yields biased synthetic outcomes, we further evaluate both a first-order Markov chain and a Qwen3-4B model fine-tuned with QLoRA across three demographic training settings: the full population, young riders only, and elderly riders only. Results show that models trained on majority-dominated populations systematically misrepresent elderly mobility behavior, particularly for spatial mobility metrics. The Markov model trained on the full population overestimates elderly step length by 4.5\% and dwell time by 8.9\%, whereas the elderly-specific model achieves substantially lower errors across most metrics. Comparisons between the Markov and LLM-based frameworks further show that higher-capability models do not necessarily improve subgroup-level fidelity under limited demographic data. These findings underscore the importance of demographic representation in mobility modeling and its downstream applications for underrepresented populations.

\keywords{Trajectory Generation \and Elderly Mobility \and Algorithmic Bias \and CitiBike \and Data Inclusivity}
\end{abstract}

%% file: 1_introduction.tex
\section{Introduction}

The integration of smart technologies into urban infrastructure has reshaped transportation planning and public health, with trajectory data mining at the core of this transformation~\cite{zheng2015trajectory}. As global populations age, ensuring that smart-city systems serve all demographic groups equitably has become a pressing concern in social computing~\cite{nilforoshan2023human}. Yet the datasets underpinning these systems rarely carry granular demographic annotations, rendering the mobility patterns of older adults largely invisible to standard analytical pipelines~\cite{feng2017mobility,demontjoye2013unique,agostini2024bayesian}. Recent large-scale analyses confirm systematic demographic disparities in human mobility data~\cite{xu2023demographic,qi2024unequal}, and a growing body of work calls for algorithmic fairness in urban computing~\cite{yan2023fairness,mehrabi2021survey}.

In age-friendly urban planning~\cite{chen2024generation}, capturing the distinct travel patterns of older adults is critical: their mobility is shaped by physical accessibility, safety concerns, and different daily activity schedules~\cite{huang2023elderly,zhang2023activity}. Mobility models trained on broad, unannotated datasets that are disproportionately representative of younger, working-age populations inherently encode demographic biases~\cite{sun2023bias}. 

When synthetic mobility data derived from majority populations informs downstream decisions such as facility allocation or public transit routing, the mobility patterns of elderly riders may become systematically underrepresented, potentially undermining initiatives such as the 15-minute city vision of localized and accessible neighborhoods~\cite{moreno2021introducing}.

A critical barrier to studying this problem is the scarcity of mobility datasets with reliable demographic labels. The Citi Bike bike-sharing system historically collected riders’ birth year and gender at account registration, providing one of the few large-scale mobility sources with age metadata~\cite{citibikeNYC}. However, citing privacy concerns, Citi Bike removed demographic attributes from public trip data beginning in 2021. 

To examine how demographic underrepresentation propagates bias into downstream mobility modeling, this work uses synthetic trajectory generation as a controlled testbed. We construct an experimental framework comparing a first-order Markov chain and a Large Language Model(LLM) under identical demographic training conditions. We focus on the 2016-2020 JC Citi Bike dataset and construct a controlled experimental design that isolates the effect of demographic composition. The main contributions are as follows: 

\begin{enumerate}
    \item \textbf{Age-Specific Mobility Characterization:} Through comparative analysis of elderly (age $\ge 65$) and young (age 18--35) rider groups in the 2016--2020 Jersey City Citi Bike, we show that the elderly exhibit a structurally distinct mobility signature: spatially confined (radius of gyration of 958~m versus 1,189~m), behaviorally routine (entropy 1.82 versus 4.15), and temporally offset from commuting peaks.

\item \textbf{Controlled Multi-Model Experiment:} We evaluate both a first-order Markov chain and a Qwen3-4B model fine-tuned with QLoRA under three demographic training settings (full population, young-only, and elderly-only). This controlled design isolates the effect of demographic composition on downstream mobility fidelity while enabling direct comparison between structured probabilistic and higher-capacity sequence modeling approaches.

    \item \textbf{Effects of Demographic Composition on Downstream Mobility Fidelity:} Our results quantify how demographic imbalance in training data propagates bias into downstream mobility modeling. Models trained on majority-dominated populations systematically overestimate elderly mobility characteristics, while elderly-specific training achieves substantially closer agreement with empirical elderly mobility patterns.

\item \textbf{Reproducible Evaluation Framework:} We propose a reproducible evaluation pipeline spanning data preprocessing, mobility metric construction, synthetic trajectory generation, and subgroup-specific evaluation. The framework provides a benchmark for evaluating demographic-specific fidelity in downstream mobility modeling.

\end{enumerate}

%% file: 2_related_work.tex
\section{Related Work}

\subsection{Human Mobility and Trajectory Generation}
Understanding human mobility is foundational to urban computing~\cite{zheng2015trajectory}. Early seminal work by Gonz{\'a}lez et al.~\cite{gonzalez2008understanding} and Song et al.~\cite{song2010limits} utilized mobile phone data to demonstrate that human movement is highly regular and predictable, exhibiting bounded spatial variance. Alessandretti et al.~\cite{alessandretti2018scaling} further revealed that human mobility operates across characteristic spatial and temporal scales, with individual trajectories exhibiting stable, idiosyncratic patterns.

Building upon these foundations, trajectory generation has evolved from statistical modeling to sophisticated deep learning techniques. Traditional approaches, such as grid-based Markov models, simulate movement via transition matrices, while data-driven routine models synthesize trajectories by learning historical motifs~\cite{pappalardo2018data}. Recently, deep generative models, including Generative Adversarial Networks (GANs)~\cite{jiang2023continuous,rao2020lstm}, sequence-to-sequence architectures, and spatial-temporal graph neural networks~\cite{wang2024spatiotemporal}, have been employed to capture complex spatial-temporal dependencies. The emergence of diffusion probabilistic models has further advanced the field: DiffTraj~\cite{zhu2023difftraj} and related methods~\cite{feng2023spatial} leverage iterative denoising to generate high-fidelity GPS trajectories with improved sample quality and diversity. Most recently,  LLM-based approaches such as MobilityGPT~\cite{wang2025mobility} have been proposed, treating trajectories as token sequences for autoregressive generation. Several comprehensive surveys have catalogued this rapidly expanding landscape~\cite{cao2023mobility,zhou2024trajectory}. However, the majority of these models are optimized for general population accuracy, often overlooking the nuanced mobility patterns of underrepresented demographic subgroups~\cite{sun2023bias}.

\subsection{Trajectory Evaluation and Privacy}
Evaluation of synthetic trajectory quality presents a multifaceted challenge. Classical approaches rely on statistical distribution matching (e.g., comparing step length, speed, radius of gyration), while sequence alignment metrics such as Dynamic Time Warping~\cite{sakoe1978dynamic} and Fr{\'e}chet distance~\cite{alt1995computing} assess structural fidelity. Recent work by Lucas et al.~\cite{lucas2023evaluating} proposed a comprehensive quality framework spanning statistical, spatial, and temporal dimensions, while Wang et al.~\cite{wang2023metrics} systematically reviewed trajectory similarity metrics in the era of deep learning.

Alongside quality concerns, privacy-preserving trajectory generation has received growing attention. Cunningham et al.~\cite{cunningham2021privacy} established foundational methods for synthetic location data with formal privacy guarantees. Chen et al.~\cite{chen2023differential} advanced differentially private trajectory generation that balances utility and privacy, while Van de Ven et al.~\cite{van2023trajectory} critically examined the limitations of generative models for trajectory privacy. Due to privacy concerns, demographic attributes such as birth year and gender were no longer publicly available in Citi Bike trip data beginning in 2018, leaving fewer demographic signals available for evaluating algorithmic fairness on this platform. Our work quantifies how the absence of such signals can reduce the fidelity of mobility models and their downstream outputs for older adults.

\subsection{Social Computing for Age-Friendly Cities}
The intersection of social computing and urban gerontology highlights the necessity of age-friendly urban environments~\cite{who2007global}. Research indicates that older adults often face mobility constraints related to physical capabilities and access to public transportation, contributing to a strong reliance on localized, neighborhood-centric travel~\cite{feng2017mobility,hjorthol2010mobility}. Recent large-scale studies leveraging mobile phone data~\cite{huang2023elderly} and GPS tracking~\cite{zhang2023activity} have provided quantitative evidence that older adults exhibit significantly smaller activity spaces, lower mobility entropy, and distinct temporal patterns compared to younger demographics.

Despite the growing emphasis on inclusive cities~\cite{chen2024generation} and the 15-minute city paradigm~\cite{moreno2021introducing}, privacy concern~\cite{demontjoye2013unique} and the limited availability of age annotations in public mobility datasets have constrained data-driven research on older adults' mobility. Recent analyses of demographic disparities in mobility data~\cite{xu2023demographic,qi2024unequal}, together with emerging work on algorithmic fairness in urban computing~\cite{yan2023fairness}, underscore the importance of addressing representation gaps. Our work bridges this gap by leveraging the 2016--2020 Jersey City subset of the Citi Bike System Data, which contains publicly available age-annotated trip records, to systematically evaluate how training-data composition affects the fidelity of mobility models for underrepresented populations.

%% file: 3_analysis.tex
\section{Characterizing Elderly Mobility Pattern}

Before evaluating how demographic composition biases downstream mobility modeling, we first establish whether elderly and young riders exhibit meaningfully different mobility patterns in the first place. This section processes the Citi Bike trip records into mobility trajectories, defines the spatial and temporal metrics used throughout the paper, and applies them to compare the two groups. The resulting characterization confirms that elderly riders follow a structurally distinct mobility signature, setting the stage for the controlled experiments in Section~4.

\subsection{Data Source and Preprocessing}

Citi Bike is a bike-sharing system operating in Jersey City and New York City. It provides trip-level records containing origin and destination stations, station coordinates, trip duration, user type, and, for records prior to 2021, rider birth year and gender. Using rider birth year, we classify 1,638,153 Jersey City trips recorded between 2016 and 2020 into age groups over a 60-month period.

After removing trips with Haversine distance greater than 10 km or computed speed exceeding 10 m/s, 93.3\% of records were retained, yielding a full-population dataset containing 1,528,325 trips from 6,102 bikes. Using rider birth year, we define elderly (65+) and young (18--35) subsets from the filtered data. The elderly subset contains 14,394 trips from 2,007 unique bikes, representing 0.9\% of all filtered trips, whereas the young subset contains 783,558 trips from 5,832 bikes, representing 51.3\%.
Trips associated with the same bike are ordered chronologically to construct mobility trajectories and compute dwell times.

\subsection{Descriptive Statistics}

Table~\ref{tab:dataset_stats} summarizes the post-filtering descriptive statistics of the JC Citi Bike 2016--2020 dataset stratified by age group.

\begin{table}[htbp]
\centering
\caption{Descriptive Statistics of the JC Citi Bike 2016--2020 dataset by age group (post-filtering).}
\label{tab:dataset_stats}
\begin{tabular}{lccccc}
\toprule
\textbf{Group} & \textbf{Trips} & \textbf{Trajectories} & \textbf{Trips/Traj.} & \textbf{Unique Stations} & \textbf{Age} \\
\midrule
Elderly (65+) & 14,394  & 2,007  & 7.2    & 50  & 65--80 \\
Young (18--35) & 783,558 & 5,832  & 134.4  & 55  & 18--35 \\
All Riders & 1,528,325   & 6,102  & 250.5  & 55  & 16--80 \\
\bottomrule
\end{tabular}
\end{table}

The elderly group is substantially smaller than the young group, accounting for 0.9\% of all filtered trips compared to 51.3\% for young riders. Elderly riders also exhibit shorter trajectories on average (7.2 vs.\ 134.4 trips per trajectory) and interact with fewer unique stations (50 vs.\ 55), indicating lower overall mobility intensity and spatial coverage.

Gender and subscription patterns are broadly similar across the two groups. Elderly riders are composed of 70\% male and 28\% female users (2\% unspecified), compared to 75\% male and 25\% female among young riders. Both groups consist almost entirely of annual Subscribers (99\%).

\subsection{Evaluation Metrics}

We evaluate synthetic mobility fidelity using spatial and temporal metrics that characterize mobility range, visitation regularity, and travel dynamics. Let a trajectory consist of $n$ trips for a given bike, where each trip $i$ contains origin and destination coordinates together with departure and arrival timestamps.

\subsubsection{Spatial Metrics}

\begin{itemize}
    \item \textbf{Step Length:} The Haversine distance between trip origin and destination stations, averaged across all trips within a group.

    \item \textbf{Radius of Gyration (RoG):} A measure of spatial dispersion computed relative to the trajectory center of mass $(\overline{\text{lat}}, \overline{\text{lon}})$:
    \begin{equation}
        r_g = \sqrt{\frac{1}{n} \sum_{i=1}^{n} d\big((\text{lat}_i^e,\text{lon}_i^e),\ (\overline{\text{lat}}, \overline{\text{lon}})\big)^2}
    \end{equation}

    \item \textbf{Mobility Entropy:} The Shannon entropy of station visitation frequencies:
    \begin{equation}
        S = - \sum_{j} p_j \log_2(p_j)
    \end{equation}
    where $p_j$ denotes the fraction of trips ending at station $j$. Lower entropy indicates more regular and predictable mobility patterns.
\end{itemize}

\subsubsection{Temporal Metrics}

\begin{itemize}
    \item \textbf{Speed:} The ratio of trip distance to trip duration.

    \item \textbf{Dwell Time:} The idle interval $t_{i+1}^s - t_i^e$ between two consecutive trips on the same bike. Dwell times are restricted to $[0,86400)$ seconds to exclude overnight or long-term inactivity.

    \item \textbf{Intra-day Temporal Distribution:}  The normalized hourly distribution of trip start times aggregated over 24-hour bins.
\end{itemize}

\subsection{Demographic Mobility Characterization}

\begin{figure}[!t]
\centering
\includegraphics[width=\textwidth]{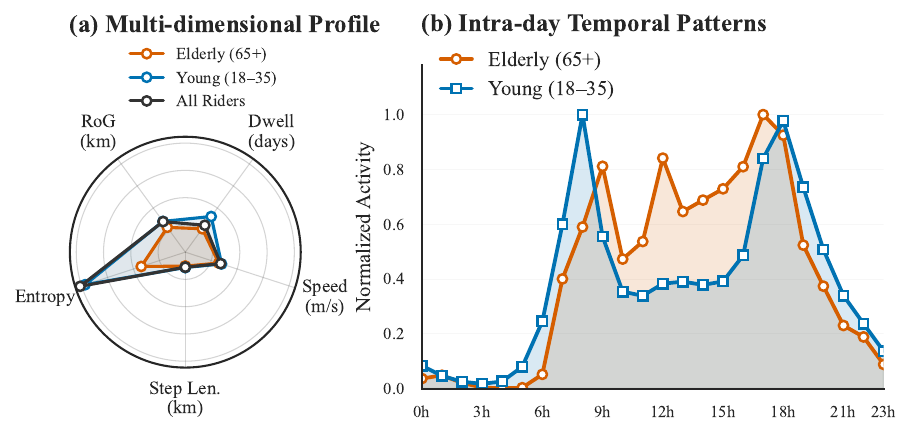}
\caption{Demographic mobility signatures of elderly (65+) and young (18--35) riders derived from JC CitiBike 2016--2020 data.}
\label{fig:mobility_signatures}
\end{figure}

Figure~\ref{fig:mobility_signatures} summarizes demographic differences in mobility structure and temporal activity patterns between elderly and young riders.

\textbf{Spatial Mobility Differences.} Figure~\ref{fig:mobility_signatures}(a) compares mobility metrics between elderly and young riders. Step lengths and travel speeds remain broadly similar across the two groups (1,004 m vs.\ 1,102 m; 2.41 m/s vs.\ 2.61 m/s). In contrast, elderly riders exhibit lower spatial dispersion and mobility diversity, with a smaller radius of gyration (958 m vs.\ 1,189 m) and substantially lower mobility entropy (1.82 vs.\ 4.15).

The aggregate all-rider entropy (4.36) exceeds that of the young group alone, indicating that elderly riders contribute additional station visitation diversity beyond the dominant young-rider mobility pattern. This difference is further reflected in the station-set Jaccard similarity between the two groups, which is 0.521.

\textbf{Temporal Activity Patterns.} Figure~\ref{fig:mobility_signatures}(b) compares intra-day mobility activity between the two groups. Young riders exhibit pronounced morning and evening activity peaks around 08:00 and 17:00--18:00, whereas elderly riders maintain more evenly distributed daytime activity concentrated between 09:00 and 17:00. Elderly riders also exhibit shorter dwell intervals between trips (18,329 s vs.\ 27,928 s), indicating distinct temporal usage patterns between the two groups.

%% file: 4_generative_experiments.tex
\section{Demographic Bias in Mobility Modeling}

Section~3 established that elderly and young riders follow structurally distinct mobility patterns. These differences carry direct implications for downstream urban computing tasks, such as facility allocation and transit routing, that depend on faithful mobility representations. A foundational component shared by these tasks is trajectory generation: synthetic mobility data are used to augment sparse observations, power urban simulations, and enable privacy-preserving data sharing. As LLMs emerge as the emerging paradigm for trajectory-based mobility analysis, understanding whether demographic bias propagates through generative pipelines becomes increasingly urgent. This section examines the question through a controlled experiment in which a first-order Markov chain and a Qwen3-4B model (QLoRA fine-tuned) are each trained under elderly-only, young-only, and full-population settings, and evaluated against real elderly trajectories using the metrics defined in Section~3.3.

\subsection{Experiment Design}
\label{sec:controlled_experiment}

To evaluate how training-data demographics affect synthetic elderly mobility fidelity, we construct a controlled generative experiment using three training populations: elderly-only (65+), young-only (18–35), and the full rider population. For both the Markov and LLM paradigms, separate models are trained on each population:

\begin{itemize}
    \item \textbf{$\mathcal{M}_{\text{elderly}}$ / $\mathcal{L}_{\text{elderly}}$:} Trained exclusively on elderly trips (65+).
    \item \textbf{$\mathcal{M}_{\text{young}}$ / $\mathcal{L}_{\text{young}}$:} Trained exclusively on young trips (18--35).
    \item \textbf{$\mathcal{M}_{\text{all}}$ / $\mathcal{L}_{\text{all}}$:} Trained on the full filtered dataset (all age groups).
\end{itemize}

Each model generates 2,007 synthetic trajectories to match the empirical elderly trajectory count. Synthetic trajectory lengths are sampled from the real elderly trajectory-length distribution (mean 7.2 trips per trajectory). All generated datasets are evaluated against the real elderly trajectories using the identical mobility evaluation pipeline.

Comparisons across training populations assess how demographic composition affects the fidelity of synthetic elderly mobility trajectories, while comparisons between the Markov and LLM paradigms examine the role of generative framework choice under severe minority-data sparsity.

\subsection{Markov Chain Generative Model}

We implement a first-order Markov chain model to generate synthetic mobility trajectories from empirical trip sequences. The model estimates station-to-station transition probabilities $P(s_j \mid s_i)$, start-station probabilities $P_{\text{start}}(s_i)$, and empirical distributions for dwell time, departure hour, and trajectory length. Trip durations for each station pair $(i,j)$ are modeled using Gaussian distributions parameterized as $\mathcal{N}(\mu_{ij}, \sigma_{ij})$.

Synthetic trajectories are generated by first sampling a start station and trajectory length, followed by iterative sampling of destination stations conditioned on the current station. Each transition is assigned a duration sampled from the corresponding transition-specific Gaussian distribution, truncated to a minimum of 60 seconds. Dwell intervals between consecutive trips are sampled from the empirical dwell-time distribution and capped at 24 hours, while departure hours are sampled from the empirical hour-of-day distribution with randomized minutes and seconds. For stations without outgoing transitions in the training data, destination sampling falls back to the start-station distribution $P_{\text{start}}(s_i)$. Synthetic datasets are generated by repeating this procedure until the target trajectory count is reached.

\subsection{LLM-based Generative Model}

We also implement a Qwen3-4B \cite{qwen3} LLM fine-tuned using 4-bit NormalFloat QLoRA \cite{qlora} to evaluate a higher-capacity generative framework under sparse demographic data. LoRA adapters are applied to all linear layers with rank $r=16$, $\alpha=32$, and dropout 0.05, yielding approximately 100M trainable parameters.

\subsubsection{Trajectory Serialization and Fine-Tuning. }

Each trajectory is represented as a whitespace-separated sequence of station IDs, such as \texttt{3195 3205 3186 \dots}. Trajectories are tokenized using the Qwen3 tokenizer and truncated to 768 tokens. Separate LoRA adapters are fine-tuned for 3 epochs using the elderly-only, young-only, and full-population training corpora described in Section~\ref{sec:controlled_experiment}, with batch size 4 and learning rate $2\times10^{-4}$. Training corpus sizes are summarized in Table~\ref{tab:llm_corpora}.

\begin{table}[htbp]
\centering
\caption{LLM training corpora (Jersey City 2016--2020, after filtering).}
\label{tab:llm_corpora}
\begin{tabular}{lrr}
\toprule
\textbf{Training Corpus} & \textbf{Trajectories} & \textbf{Trips} \\
\midrule
Elderly (65+)              & 1,631                & 13,928 \\
Young (18--35)             & 2,897                & 469,801 \\
All Riders                & 2,984                & 667,013 \\
\bottomrule
\end{tabular}
\end{table}

\subsubsection{Generation Procedure.}

The fine-tuned model generates trajectories via batched autoregressive decoding (batch size 32) with temperature 0.7, top-$p$ 0.9, and top-$k$ 50, producing up to 128 new tokens per trajectory. Decoding is seeded with a start-of-sequence token. The generated token stream is parsed back into station ID integers by concatenating consecutive digit tokens; non-numeric tokens and station IDs not present in the training vocabulary are discarded. Each model generates 2,007 synthetic trajectories to match the real elderly trajectory count.

\subsubsection{Temporal Synthesis.}

Since the LLM operates on station sequences only and does not model continuous temporal quantities, we synthesize temporal attributes post hoc using fixed heuristics. Trip duration is set to trip distance divided by 4.17~m/s (the real elderly mean speed), and inter-trip dwell time is sampled from an exponential distribution with mean 600~s. These heuristics are identical across all three LLM-trained models.

\subsection{Markov-based Generation Results}

\begin{figure}[!t]
\centering
\includegraphics[width=\textwidth]{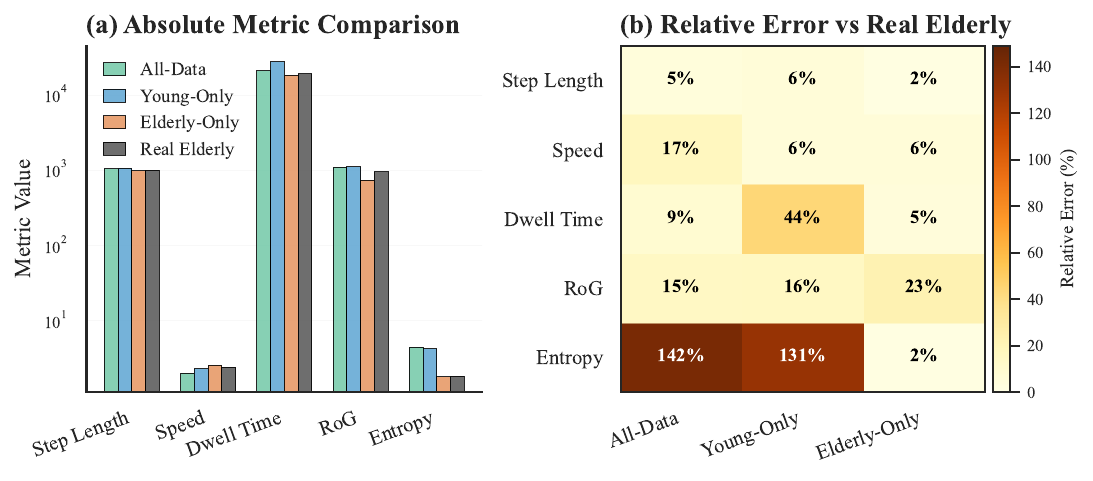}
\caption{Synthetic trajectory evaluation under different training populations. (a) Absolute comparison of mobility metrics across Real Elderly and synthetic trajectories generated by $\mathcal{M}_{\text{full}}$, $\mathcal{M}_{\text{young}}$, and $\mathcal{M}_{\text{elderly}}$ (log scale). (b) Relative error (\%) of each synthetic method with respect to Real Elderly across the five mobility metrics.}
\label{fig:minority_imperative}
\end{figure}

We evaluate how training-data composition affects the fidelity of synthetic elderly mobility trajectories using three Markov models trained on different demographic populations: $\mathcal{M}_{\text{full}}$, $\mathcal{M}_{\text{young}}$, and $\mathcal{M}_{\text{elderly}}$. Each model generates 2,007 synthetic trajectories matching the empirical elderly trajectory count and length distribution. Figure~\ref{fig:minority_imperative} and Table~\ref{tab:gen_results} summarize the resulting mobility metrics relative to the real elderly baseline.

\begin{table}[!t]
\centering
\caption{Comparison of real elderly metrics with synthetic trajectories generated by Markov models trained on different demographic compositions.}
\label{tab:gen_results}
\small
\begin{tabular}{lcccc}
\toprule
\textbf{Metric} & \textbf{Real Elderly} & \textbf{$\mathcal{M}_{\text{full}}$} & \textbf{$\mathcal{M}_{\text{young}}$} & \textbf{$\mathcal{M}_{\text{elderly}}$} \\
\midrule
Step Length (m) & 1,004.3 & 1,049.5 (+4.5\%) & 1,067.7 (+6.3\%) & \textbf{984.2 ($-$2.0\%)} \\
Speed (m/s) & 2.41  & 2.00 ($-$17.0\%)  & 2.27 ($-$5.9\%)      & \textbf{2.56 (+6.2\%)} \\
Dwell Time (s) & 19,338 & 21,064 (+8.9\%) & 27,893 (+44.3\%)   & \textbf{18,392 ($-$4.9\%)} \\
RoG (m) & 957.8 & 1,101.1 (+15.0\%) & 1,106.9 (+15.6\%)     & 738.8 ($-$22.9\%) \\
Entropy & 1.82 & 4.41 (+141.9\%) & 4.21 (+130.8\%) & \textbf{1.79 ($-$1.8\%)} \\
\bottomrule
\end{tabular}
\end{table}

$\mathcal{M}_{\text{young}}$ exhibits substantial deviations from the real elderly baseline, particularly for Dwell Time (+44.3\%), Entropy (+130.8\%), and RoG (+15.6\%), indicating that mobility patterns learned exclusively from young riders do not transfer well to elderly mobility behavior.

Including all demographic groups in the training data improves temporal fidelity. Compared with $\mathcal{M}_{\text{young}}$, $\mathcal{M}_{\text{full}}$ reduces the Dwell Time error from 44.3\% to 8.9\%. However, spatial metrics remain poorly reproduced, with Entropy and RoG errors remaining comparable to the young-only model. This suggests that aggregate transition patterns remain dominated by the majority population.

Among the three variants, $\mathcal{M}_{\text{elderly}}$ achieves the closest agreement with the real elderly baseline for Step Length, Dwell Time, and Entropy, with errors below 5\%. However, it underestimates RoG by 22.9\%, indicating reduced spatial coverage in the generated trajectories. This contraction likely reflects the sparsity of elderly transition observations in the training data.

Despite the substantially larger training corpus used in the full-population setting, the spatial fidelity gap remains unresolved. Together, these results indicate that demographic inclusion alone is insufficient to recover minority-group spatial mobility structure when observational coverage is sparse.

\subsection{LLM-based Generation Results}

We repeat the synthetic trajectory experiment using Qwen3-4B with QLoRA fine-tuning. Three adapters ($\mathcal{L}_{\text{elderly}}$, $\mathcal{L}_{\text{young}}$, $\mathcal{L}_{\text{all}}$) are trained on the same demographic splits used in the Markov experiments, with each model generating 2,007 synthetic trajectories. Table~\ref{tab:llm_results} summarizes the evaluation against the real elderly baseline.

\begin{table}[htbp]
\centering
\caption{LLM-generated synthetic trajectories vs.\ Real Elderly. Bold indicates best among the three LLM variants.}
\label{tab:llm_results}
\small
\begin{tabular}{lcccc}
\toprule
\textbf{Metric} & \textbf{Real Elderly} & \textbf{$\mathcal{L}_{\text{elderly}}$} & \textbf{$\mathcal{L}_{\text{young}}$} & \textbf{$\mathcal{L}_{\text{all}}$} \\
\midrule
Step Length (m) & 1,004.3 & 1,189.2 (+18.4\%) & 967.3 ($-$3.7\%) & \textbf{878.8 ($-$12.5\%)} \\
Speed (m/s) & 2.41  & 4.15 (+72.3\%)  & 4.15 (+72.2\%)      & 4.16 (+72.4\%) \\
Dwell Time (s) & 19,338 & 597 ($-$96.9\%) & 601 ($-$96.9\%)   & 605 ($-$96.9\%) \\
RoG (m) & 957.8 & \textbf{903.6 ($-$5.7\%)} & 846.9 ($-$11.6\%)     & 773.5 ($-$19.2\%) \\
Entropy & 1.82 & \textbf{2.72 (+49.1\%)} & 3.06 (+67.6\%) & 2.86 (+57.0\%) \\
\bottomrule
\end{tabular}
\end{table}

Across the three LLM variants, $\mathcal{L}_{\text{elderly}}$ achieves the lowest errors for the spatial metrics RoG and Entropy. However, all LLM variants exhibit substantial errors for the temporal metrics Speed and Dwell Time. Because temporal attributes are synthesized post hoc rather than directly generated by the model, temporal patterns remain poorly aligned with the empirical elderly baseline.

Compared with the Markov models, the LLM-based models generally achieve lower fidelity across most metrics. $\mathcal{L}_{\text{elderly}}$ underperforms $\mathcal{M}_{\text{elderly}}$ on Step Length, Speed, Dwell Time, and Entropy. The primary exception is RoG, where the LLM achieves a smaller error ($-$5.7\%) than the Markov model ($-$22.9\%). This suggests that autoregressive sequence generation partially alleviates the spatial contraction caused by sparse transition observations. However, this improvement in spatial coverage comes at the cost of reduced behavioral regularity, as the LLM models tend to overestimate mobility diversity, particularly for Entropy.

Overall, the results suggest that higher-capacity sequence models do not necessarily improve minority-group mobility fidelity when demographic training data remain limited.

%% file: 5_conclusions_and_policy_implications.tex
\section{Conclusions and Policy Implications}

\subsection{Conclusions}

This study examined how demographic underrepresentation biases mobility models, using synthetic trajectory generation as a controlled testbed. Elderly and young riders exhibit structurally distinct mobility patterns across spatial and temporal dimensions. Controlled experiments show that models trained on majority-dominated populations systematically misrepresent elderly mobility behavior, while elderly-specific training achieves closer agreement with empirical patterns, although spatial coverage remains constrained by data sparsity. Comparisons between Markov and LLM-based generation further reveal that higher-capacity models do not necessarily improve subgroup-level fidelity under limited demographic data.

\subsection{Policy Implications}

The results suggest that aggregate mobility metrics may mask substantial behavioral differences across demographic groups. Models trained primarily on majority populations can misrepresent minority mobility behavior even when overall performance appears acceptable. Incorporating demographic-specific evaluation into mobility analysis may therefore improve the reliability of data-driven transportation planning and policy decisions.

The empirical mobility patterns observed among elderly riders also indicate that bike-sharing systems may serve older populations differently from younger users. Elderly riders exhibited more localized spatial activity, lower mobility diversity, and distinct temporal usage patterns, suggesting that age-inclusive mobility infrastructure and improved station accessibility may help broaden participation among older adults.

More broadly, as data-driven mobility models become increasingly integrated into urban analytics and smart-city systems, evaluation based solely on aggregate performance metrics may overlook systematic demographic-specific errors. Whether the downstream task is generation, prediction, or resource allocation, demographic-aware validation procedures can help improve the robustness and representativeness of mobility models used in urban decision-making.

\subsection{Limitations and Future Work}

This study is subject to several limitations. First, the analysis is restricted to the Jersey City Citi Bike system, and future work should evaluate whether similar subgroup-specific mobility patterns emerge across larger and more diverse urban settings. Second, demographic analysis is limited to age group. Extending subgroup-level mobility evaluation to additional demographic dimensions and downstream tasks remains an important direction for future research. Finally, this study captures mobility patterns within a specific historical period (2016--2020), while urban mobility systems and demographic travel behaviors may evolve gradually over longer time horizons. Future research should examine whether subgroup-specific mobility disparities persist under long-term changes in urban mobility systems.